\documentclass[11pt,letterpaper]{article}
\usepackage{emnlp2016}
\usepackage{times}
\usepackage{latexsym}
\usepackage{graphicx}
\usepackage{booktabs}
\usepackage{amsmath}
\usepackage{amssymb}
\usepackage{bbm}
\usepackage{xspace}
\usepackage{algorithm}
\usepackage{calrsfs}
\usepackage[noend]{algpseudocode}
\usepackage{tabularx}
\usepackage{pbox}
\usepackage{microtype} 
\usepackage{hyperref}
\usepackage[usenames,dvipsnames,svgnames,table]{xcolor}
\usepackage{subfig}

\emnlpfinalcopy

\def\emnlppaperid{159}

 \makeatletter
 \DeclareRobustCommand\onedot{\futurelet\@let@token\@onedot}
 \def\@onedot{\ifx\@let@token.\else.\null\fi\xspace}
 \def\eg{e.g\onedot} 
 \def\ie{i.e\onedot}

 \makeatother

\DeclareRobustCommand{\Figref}[1]{Figure~\ref{#1}}

\DeclareRobustCommand{\secref}[1]{Sec.~\ref{#1}}
\DeclareRobustCommand{\Secref}[1]{Sec.~\ref{#1}}

\DeclareRobustCommand{\Eqnref}[1]{Equation~(\ref{#1})}

\newcommand{\myparagraph}[1]{\paragraph{#1}}

\title{Multimodal Compact Bilinear Pooling\\for Visual Question Answering and Visual Grounding}

\newcommand{\authSpace}{\ \ \ \ \ \ }
\newcommand\blfootnote[1]{%
  \begingroup
  \renewcommand\thefootnote{}\footnote{#1}%
  \addtocounter{footnote}{-1}%
  \endgroup
}

\author{
 Akira Fukui*$^{1,2}$ \authSpace Dong Huk Park*$^{1}$ \authSpace Daylen Yang*$^{1}$ \\ {\bf Anna Rohrbach*}$^{1,3}$ \authSpace {\bf Trevor Darrell}$^{1}$ \authSpace {\bf Marcus Rohrbach}$^{1}$\\
$^{1}$UC Berkeley EECS,  CA, United States\\
$^{2}$Sony Corp., Tokyo, Japan\\
$^{3}$Max Planck Institute for Informatics, Saarbr{\"u}cken, Germany\\
}

\newcommand{\twocmidrule}{\cmidrule(lr){1-1}\cmidrule(lr){2-2}}

\date{}

\begin{document}

\maketitle
\blfootnote{* indicates equal contribution}

\begin{abstract}
Modeling textual or visual information with vector representations trained from large language or visual datasets has been successfully explored in recent years. However, tasks such as  visual question answering require combining these vector representations with each other. Approaches to multimodal pooling include element-wise product or sum, as well as concatenation of the visual and textual representations. We hypothesize that these methods are not as expressive as an outer product of the visual and textual vectors. As the outer product is typically infeasible due to its high dimensionality, we instead propose utilizing Multimodal Compact Bilinear pooling (MCB) to efficiently and expressively combine multimodal features. We extensively evaluate MCB on the visual question answering and grounding tasks. We consistently show the benefit of  MCB over ablations without MCB. For visual question answering, we present an architecture which uses MCB twice, once for predicting attention over spatial features and again to combine the attended representation with the question representation. This model outperforms the state-of-the-art on the Visual7W dataset and the VQA challenge.
\end{abstract}

\section{Introduction}
\begin{figure}[t]
\includegraphics[width=.475\textwidth]{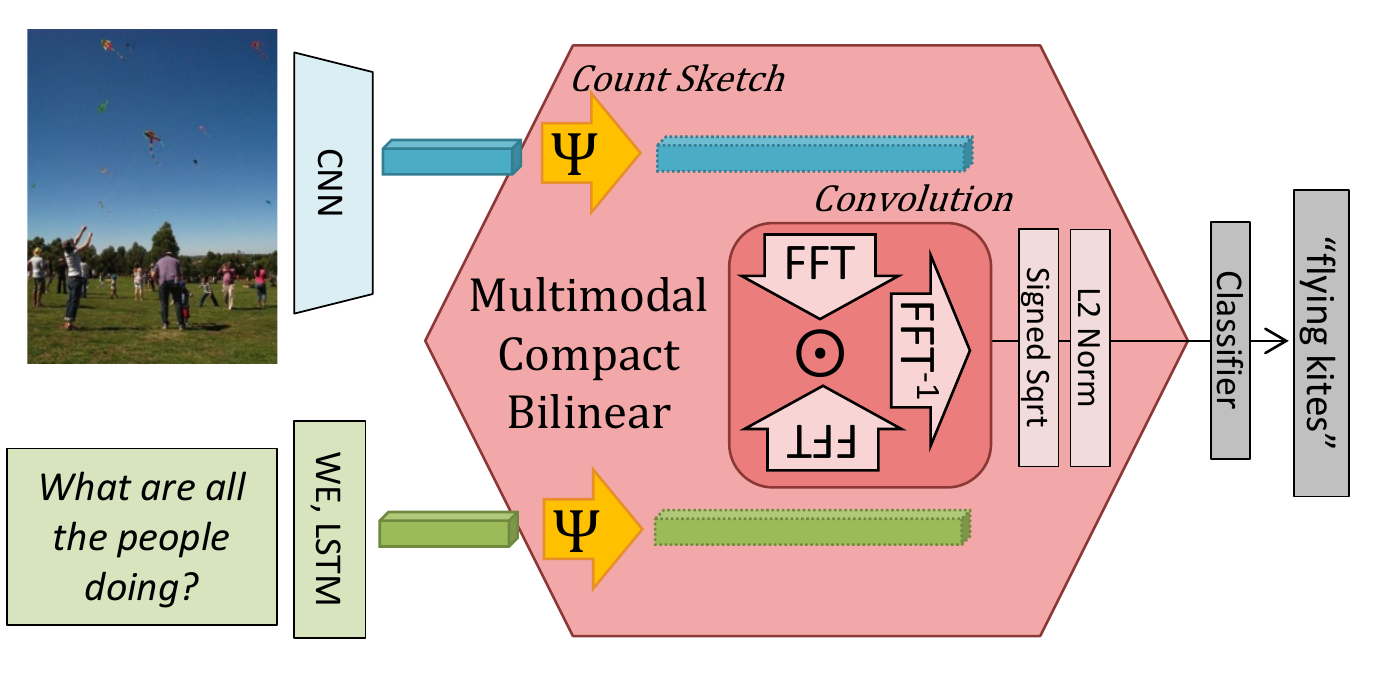}
\vspace{-0.5cm}
\caption{Multimodal Compact Bilinear Pooling for visual question answering.}
\label{fig:teaser}
\end{figure}

Representation learning for text and images has been extensively studied in recent years. Recurrent neural networks (RNNs) are often used to represent sentences or phrases \cite{sutskever14nips,kiros15nips}, and convolutional neural networks (CNNs) have shown to work best to represent images \cite{donahue14icml,he2015deep}. For tasks such as visual question answering (VQA) and visual grounding, most approaches require joining the representation of both modalities. %
For combining the two vector representations (multimodal pooling),
 current approaches in VQA or grounding rely on concatenating vectors or applying element-wise sum or product. While this generates a joint representation, it might not be expressive enough to fully capture the complex associations between the two different modalities. 
    
In this paper, we propose to rely on Multimodal Compact Bilinear pooling (MCB)   to get a joint representation. Bilinear pooling  computes the outer product between two vectors, which allows, in contrast to element-wise product, a multiplicative interaction between all elements of both vectors. Bilinear pooling models \cite{tenenbaum00nc} have  recently been shown to be beneficial for fine-grained classification  for vision only tasks \cite{lin15iccv}. However, given their high dimensionality ($n^2$), bilinear pooling has so far not been widely used. In this paper, we adopt the idea from \newcite{gao16cvpr} which shows how to efficiently compress bilinear pooling for a single modality.
In this work, we discuss and extensively evaluate the extension to the multimodal case for text and visual modalities.
As shown in  \Figref{fig:teaser},  Multimodal Compact  Bilinear pooling (MCB) is approximated by randomly projecting the image and text representations to a higher dimensional space (using Count Sketch \cite{charikar2002countsketch}) and then convolving both vectors efficiently by using element-wise product in Fast Fourier Transform (FFT) space.
We use MCB to predict answers for the VQA task and locations for the visual grounding task. For open-ended question answering, we present an architecture for VQA which uses MCB twice, once to predict spatial attention and the second time to predict the answer. For multiple-choice question answering we introduce a third MCB to relate the encoded answer to the question-image space.
Additionally, we discuss the benefit of attention maps and additional training data for the VQA task.
To summarize, MCB is evaluated on two tasks, four datasets, and with a diverse set of ablations and comparisons to the state-of-the-art.

\section{Related Work}

\paragraph{Multimodal pooling.}
Current approaches to multimodal pooling involve element-wise operations or vector concatenation.
In the visual question answering domain, a number of models have been proposed. Simpler models such as iBOWIMG baseline \cite{zhou2015simple} use concatenation and fully connected layers to combine the image and question modalities.
Stacked Attention Networks \cite{yang2015stacked} and Spatial Memory Networks \cite{xu15icml} use LSTMs or extract soft-attention on the image features, but ultimately use element-wise product or element-wise sum to merge modalities.
D-NMN \cite{andreas16naacl} introduced REINFORCE to dynamically create a network and use element-wise product to join attentions and element-wise sum predict answers.
Dynamic Memory Networks (DMN) \cite{xiong16dynamic} pool the image and question with element-wise product and sum, attending to part of the image and question with an Episodic Memory Module \cite{kumar15arxiv}.
DPPnet \cite{noh2015images} creates a Parameter Prediction Network which learns to predict the parameters of the second to last visual recognition layer dynamically from the question. Similar to this work, DPPnet allows multiplicative interactions between the visual and question encodings.
\newcite{lu2016hiecoatt} recently proposed a model that extracts multiple co-attentions on the image and question and combines the co-attentions in a hierarchical manner using element-wise sum, concatenation, and fully connected layers.

For the visual grounding task, \newcite{rohrbach16arxiv} propose an approach where the language phrase embedding is concatenated with the visual features in order to predict the attention weights over multiple bounding box proposals. Similarly, \newcite{hu16arxiv} concatenate phrase embeddings with visual features at different spatial locations to obtain a segmentation.

\paragraph{Bilinear pooling.} Bilinear pooling has been applied to the fine-grained visual recognition task. \newcite{lin15iccv} use two CNNs to extract features from an image and combine the resulting vectors using an outer product, which is fully connected to an output layer. \newcite{gao16cvpr} address the space and time complexity of bilinear features by viewing the bilinear transformation as a polynomial kernel. \newcite{pham2013tensorsketch} describe a method to approximate the polynomial kernel using Count Sketches and convolutions. %

\paragraph{Joint multimodal embeddings.}
In order to model similarities between two modalities, many prior works have learned joint multimodal spaces, or embeddings. Some of such embeddings are based on Canonical Correlation Analysis \cite{hardoon2004canonical} \eg \cite{gong14eccv,klein15cvpr,plummer15iccv}, linear models with ranking loss \cite{frome2013devise,karpathy15cvpr,socher2014grounded,weston2011wsabie} or non-linear deep learning models \cite{kiros2014multimodal,mao2014deep,ngiam2011multimodal}. Our multimodal compact bilinear pooling can be seen as a complementary operation that allows us to capture different interactions between two modalities more expressively than \eg concatenation. Consequently, many embedding learning approaches could benefit from incorporating such interactions.

\section{Multimodal Compact Bilinear Pooling for Visual and Textual Embeddings}
\label{sec:approach}

\begin{figure}[t]
\includegraphics[width=0.475\textwidth]{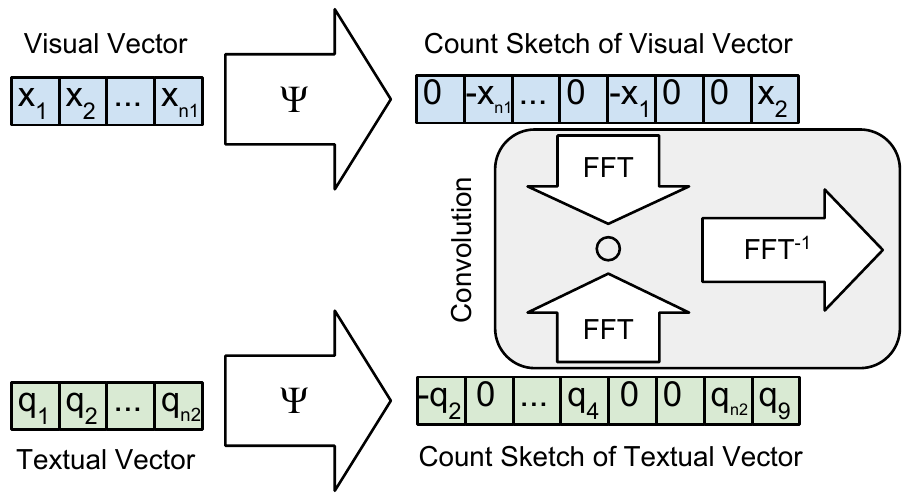}
\vspace{-0.5cm}
\caption{Multimodal Compact Bilinear Pooling (MCB)}
\label{fig:count_sketch}
\end{figure}

For the task of visual question answering (VQA) or visual grounding, we have to predict the most likely answer or location $\hat{a}$ for a given image $\textbf{x}$ and question or phrase $\textbf{q}$. This can be formulated as

\newcommand*{\argmin}{\operatornamewithlimits{argmin}\limits}
\newcommand*{\argmax}{\operatornamewithlimits{argmax}\limits}

\begin{equation}
\label{eq:problem}
\hat{a}=\argmax_{a\in A}p(a|\textbf{x},\textbf{q};\theta)
\end{equation}
with parameters $\theta$ and the set of answers or locations $A$.
For an image embedding $x = \Xi(\textbf{x})$ (\ie a CNN) and question embedding $q = \Omega(\textbf{q})$ (\ie an LSTM), we are interested in getting a good joint representation by pooling both representations. With a multimodal pooling $\Phi(x, q)$ that encodes the relationship between $x$ and $q$ well, it becomes easier to learn a classifier for \Eqnref{eq:problem}. 

In this section, we first discuss our multimodal pooling $\Phi$ for combining representations from different modalities into a single representation (\secref{sec:approach:mcb}) and then detail our architectures for VQA (\secref{sec:architecture:vqa}) and visual grounding (\secref{sec:architecture:grounding}), further explaining how we predict $\hat{a}$ with the given image representation $\Xi$ and text representation $\Omega$.

\subsection{Multimodal Compact Bilinear Pooling~(MCB)}
\label{sec:approach:mcb}

\begin{algorithm}
\small
\caption{Multimodal Compact Bilinear}\label{alg:ts}
\begin{algorithmic}[1]
\State input: $v_1 \in \mathbb{R}^{n_1}, v_2 \in \mathbb{R}^{n_2}$
\State output: $\Phi(v_1, v_2) \in \mathbb{R}^d$
\Procedure{MCB}{$v_1, v_2, n_1, n_2, d$}
\For{$k \leftarrow 1\ldots 2$}
\If {$h_k, s_k$ not initialized}
\For{$i \leftarrow 1\ldots n_k$}
\State sample $h_{k}[i]$ from $\{1,\ldots,d\}$
\State sample $s_{k}[i]$ from $\{-1,1\}$
\EndFor
\EndIf
\State $v_k' = \Psi(v_k, h_k, s_k, n_k)$
\EndFor
\State $\Phi = \text{FFT}^{-1}(\text{FFT}(v_1') \odot \text{FFT}(v_2')) $
\State \Return $\Phi$
\EndProcedure

\Procedure{$\Psi$}{$v, h, s, n$}
\State $y = [0, \ldots, 0]$
\For{$i\leftarrow 1\ldots n$}
\State $y[h[i]] = y[h[i]] + s[i] \cdot v[i]$
\EndFor
\State \Return $y$
\EndProcedure
\end{algorithmic}
\end{algorithm}
Bilinear models \cite{tenenbaum00nc} take the outer product of two vectors $x \in \mathbb{R}^{n_1}$ and $q \in \mathbb{R}^{n_2}$ and learn a model $W$ (here linear), \ie
$z=W\left[x \otimes q\right]$, 
where  $\otimes$ denotes the outer product ($xq^T$) and $\left[\ \right]$ denotes linearizing the matrix in a vector.
As discussed in the introduction, bilinear pooling is interesting because it allows all elements of both vectors to interact with each other in a multiplicative way. 
However, the high dimensional representation (\ie when $n_1$ and $n_2$ are large) leads to an infeasible number of parameters to learn in $W$. For example, we use $n_1 = n_2 = 2048$ and  $z \in \mathbb{R}^{3000}$ for VQA. $W$ thus would have 12.5 billion parameters, which leads to very high memory consumption and high computation times.

We thus need a method that projects the outer product to a lower dimensional space and also avoids computing the outer product directly. As suggested by \newcite{gao16cvpr} for a single modality, we rely on the Count Sketch projection function $\Psi$ \cite{charikar2002countsketch}, which projects a vector $v \in \mathbb{R}^n$ to $y \in \mathbb{R}^d$.
We initialize two vectors $s \in \{-1,1\}^n$  %
and $h \in \{1,...,d\}^n$:  %
$s$ contains either $1$ or $-1$ for each index, and $h$ maps each index $i$ in the input $v$ to an index $j$ in the output $y$. Both $s$ and $h$ are initialized randomly from a uniform distribution and remain constant for future invocations of count sketch. $y$ is initialized as a zero vector. For every element    $v[i]$ its destination index $j=h[i]$ is looked up using $h$, and $s[i] \cdot{} v[i] $ is added to $y[j]$. See lines 1-9 and 12-16 in Algorithm \ref{alg:ts}.

\begin{figure*}[t]
\centering
\includegraphics[width=0.8\textwidth]
{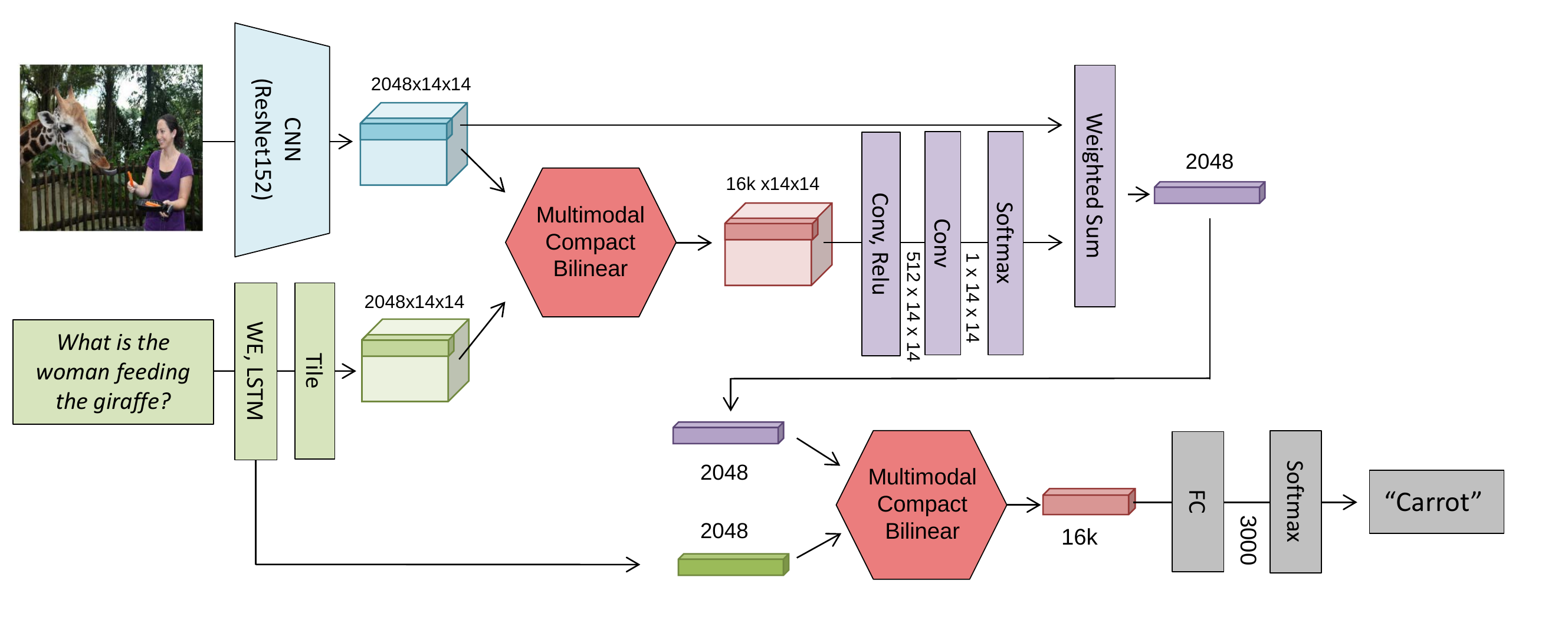}
\vspace{-0.5cm}
\caption{Our architecture for VQA: Multimodal Compact Bilinear (MCB) with Attention. Conv implies convolutional layers and FC implies fully connected layers. For details see \Secref{sec:architecture:vqa}.}
\vspace{-0.5cm}
\label{fig:model_attention}
\end{figure*}
This allows us to project the outer product to a lower dimensional space, which reduces the number of parameters in $W$. To avoid computing the outer product explicitly, \newcite{pham2013tensorsketch} showed that the count sketch of the outer product of two vectors can be expressed as convolution of both count sketches: $\Psi (x \otimes q,h,s) = \Psi(x,h,s) * \Psi(q,h,s) $, where $*$ is the convolution operator. Additionally, the convolution theorem states that convolution in the time domain is equivalent to element-wise product in the frequency domain. The convolution $x' * q'$ can be rewritten as $\text{FFT}^{-1}(\text{FFT}(x') \odot \text{FFT}(q'))$, where $\odot$ refers to element-wise product.
These ideas are summarized in Figure \ref{fig:count_sketch} and formalized in Algorithm \ref{alg:ts}, which is based on the Tensor Sketch algorithm of \newcite{pham2013tensorsketch}. We invoke the algorithm with $v_1 = x$ and $v_2 = q$. We note that this easily extends and remains efficient for more than two multi-modal inputs as the combination happens as element-wise product.

\subsection{Architectures for VQA}
\label{sec:architecture:vqa}

In VQA, the input to the model is an image and a question, and the goal is to answer the question. Our model extracts representations for the image and the question, pools the vectors using MCB, and arrives at the answer by treating the problem as a multi-class classification problem with 3,000 possible classes.

We extract image features using a 152-layer Residual Network \cite{he2015deep} that is pretrained on ImageNet data \cite{deng09imagenet}. Images are resized to $ 448 \times 448 $, and we use the output of the layer (``pool5'') before the 1000-way classifier. We then perform $L_2$ normalization on the 2048-D vector.

Input questions are first tokenized into words, and the words are one-hot encoded and passed through a learned embedding layer. The tanh nonlinearity is used after the embedding. The embedding layer is followed by a 2-layer LSTM with 1024 units in each layer. The outputs of each LSTM layer are concatenated to form a 2048-D vector.

The two vectors are then passed through MCB. The MCB is followed by an element-wise signed square-root and $L_2$ normalization. After MCB pooling, a fully connected layer connects the resulting 16,000-D multimodal representation to the 3,000 top answers.

\myparagraph{Attention.}
\label{sec:architecture:att}
To incorporate spatial information, we use soft attention on our MCB pooling method.  Explored by \cite{xu15icml} for image captioning and by \cite{xu2015ask} and \cite{yang2015stacked} for VQA, the soft attention mechanism can be easily integrated in our model.

For each spatial grid location in the visual representation (i.e. last convolutional layer of ResNet [res5c], last convolutional layer of VGG [conv5]), we use MCB pooling to merge the slice of the visual feature with the language representation. As depicted in Figure \ref{fig:model_attention}, after the pooling we use two convolutional layers to predict the attention weight for each grid location. We apply softmax to produce a normalized soft attention map.
We then take a weighted sum of the spatial vectors using the attention map to create the attended  %
visual representation. We also experiment with generating multiple attention maps to allow the model to make multiple ``glimpses'' which are concatenated before being merged with the language representation through another MCB pooling for prediction. Predicting attention maps with MCB pooling allows the model to effectively learn how to attend to salient locations based on both the visual and language representations. 

\begin{figure}[t]
\includegraphics[width=0.47\textwidth]{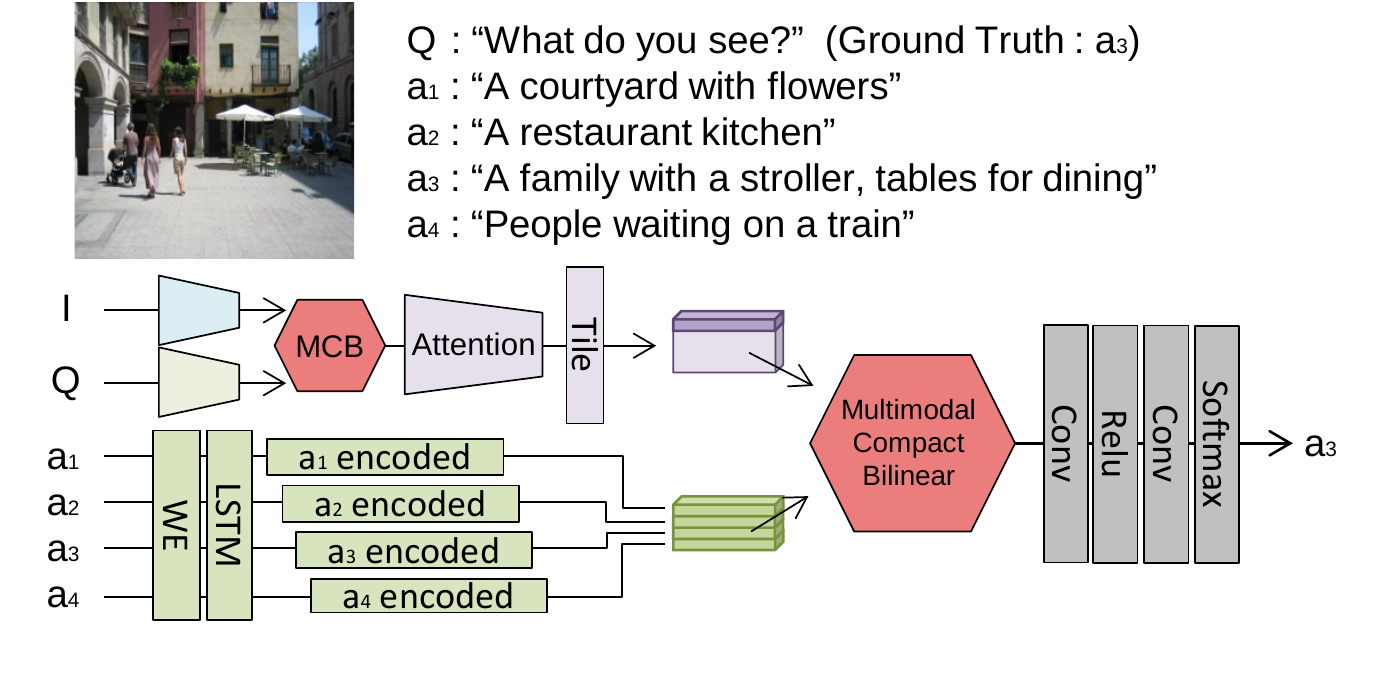}
\vspace{-0.5cm}
\caption{Our architecture for VQA: MCB with Attention and Answer Encoding}
\label{fig:model_7w}
\end{figure}

\myparagraph{Answer Encoding.}
\label{sec:architecture:visual7w}
For VQA with multiple choices, we can additionally embed the answers. We base our approach on the proposed MCB with attention. As can be seen from Figure \ref{fig:model_7w}, to deal with multiple variable-length answer choices, each choice is encoded using a word embedding and LSTM layers whose weights are shared across the candidates. In addition to using MCB with attention, we use an additional MCB pooling to merge the encoded answer choices with the multimodal representation of the original pipeline. The resulting embedding is projected to a classification vector with a dimension equal to the number of answers. 

\subsection{Architecture for Visual Grounding}
\label{sec:architecture:grounding}
\begin{figure}[t]
\includegraphics[width=0.47\textwidth]{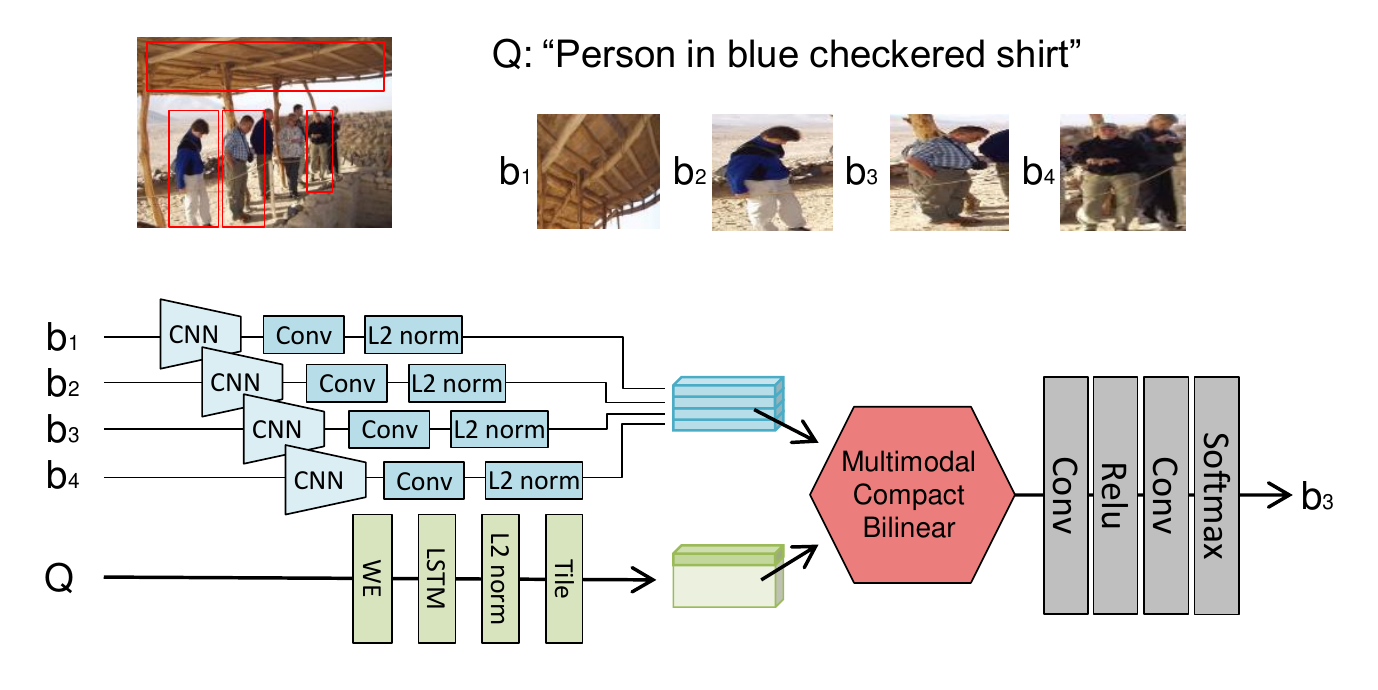}
\vspace{-0.4cm}
\caption{Our Architecture for Grounding with MCB (Sec. \ref{sec:architecture:grounding})}
\label{fig:model_grounder}
\end{figure}

We base our grounding approach on the fully-supervised version of GroundeR \cite{rohrbach16arxiv}. The overview of our model is shown in  \Figref{fig:model_grounder}. The input to the model is a query natural language phrase and an image along with multiple proposal bounding boxes. The goal is to predict a bounding box which corresponds to the query phrase. We replace the concatenation of the visual representation and the encoded phrase in GroundeR with MCB to combine both modalities. In contrast to \newcite{rohrbach16arxiv},  we include a linear embedding of the visual representation and $L_2$ normalization of both input modalities, instead of batch normalization \cite{ioffe2015batch}, which we found to be beneficial when using MCB for the grounding task.

\section{Evaluation on Visual Question Answering}
We evaluate the benefit of MCB  with a diverse set of ablations on two visual question answering datasets. %

\subsection{Datasets}
The \textbf{Visual Question Answering (VQA)} real-image dataset \cite{antol2015vqa} consists of approximately 200,000 MSCOCO images \cite{lin2014microsoft}, with 3 questions per image and 10 answers per question. There are 3 data splits: train (80K images), validation (40K images), and test (80K images). Additionally, there is a 25\% subset of test named test-dev. Accuracies for ablation experiments in this paper are reported on the test-dev data split. We use the VQA tool provided by \newcite{antol2015vqa} for evaluation. We conducted most of our experiments on the open-ended real-image task. In Table \ref{tab:vqa_state_of_art}, we also report our multiple-choice real-image scores.

The \textbf{Visual Genome} dataset \cite{krishna16arxiv} uses 108,249 images from the intersection of YFCC100M \cite{DBLP:journals/corr/ThomeeSFENPBL15} and MSCOCO. For each image, an average of 17 question-answer pairs are collected. There are 1.7 million QA pairs of the 6W question types (\textit{what, where, when, who, why}, and \textit{how}).
Compared to the VQA dataset, Visual Genome represents a more balanced distribution of the 6W question types. Moreover, the average question and answer lengths for Visual Genome are larger than the VQA dataset. To leverage the Visual Genome dataset as additional training data, we remove all the unnecessary words such as ''a'', ''the'', and ''it is'' from the answers to decrease the length of the answers and extract QA pairs whose answers are single-worded. The extracted data is filtered again based on the answer vocabulary space created from the VQA dataset, leaving us with additional 1M image-QA triplets.

The \textbf{Visual7W} dataset \cite{zhu16cvpr} is a part of the Visual Genome. Visual7W adds a 7th \textit{which} question category to accommodate visual answers, but we only evaluate the models on the Telling task which involves 6W questions. The natural language answers in Visual7W are in a multiple-choice format and each question comes with four answer candidates, with only one being the correct answer. Visual7W is composed of 47,300 images from MSCOCO and there are a total of 139,868 QA pairs. 

\begin{table}
\centering
\begin{tabular}{lc}
\toprule
 \bf Method & \bf Accuracy \\ \twocmidrule
Element-wise Sum & 56.50 \\
Concatenation & 57.49 \\
Concatenation + FC & 58.40 \\
Concatenation + FC + FC & 57.10 \\
Element-wise Product & 58.57 \\
Element-wise Product + FC & 56.44 \\
Element-wise Product + FC + FC & 57.88 \\
MCB ($2048 \times 2048 \rightarrow 16$K) & \bf 59.83 \\
\twocmidrule
Full Bilinear ($128 \times 128 \rightarrow 16$K) & 58.46 \\
MCB ($128 \times 128 \rightarrow 4$K) & 58.69 \\
\twocmidrule
Element-wise Product with VGG-19 & 55.97 \\
MCB ($d=16$K) with VGG-19 & \bf 57.05 \\
\twocmidrule
Concatenation + FC with Attention & 58.36 \\
MCB ($d=16$K) with Attention & \bf 62.50 \\
\bottomrule
\end{tabular}
\caption{Comparison of multimodal pooling methods. Models are trained on the VQA train split and tested on test-dev.}
\label{tab:comparefusion}
\end{table}
\subsection{Experimental Setup}

We use the Adam solver with $\epsilon = 0.0007$, $\beta_1 = 0.9$, $\beta_2 = 0.999$. We use dropout after the LSTM layers and in fully connected layers. For the experiments in Table \ref{tab:comparefusion} and  \ref{tab:comparedvalue}, we train on the VQA train split, validate on the VQA validation split, and report results on the VQA test-dev split. We use early stopping: if the validation score does not improve for 50,000 iterations, we stop training and evaluate the best iteration on test-dev.

For the Visual7W task, we use the same hyperparameters and training settings as in the VQA experiments. We use the splits from \cite{zhu16cvpr} to train, validate, and test our models. We also compute accuracies on this data using their evaluation code.

For VQA multiple choice, we train the open-ended models and take the argmax over the multiple choice answers at test time. For Visual7W, we use the answer encoding as described in \secref{sec:architecture:vqa}.

\begin{table}
\centering
\small
\begin{tabular}{lc}
\toprule
\bf Compact Bilinear $d$ & \bf Accuracy \\ 
\twocmidrule
1024 & 58.38 \\
2048 & 58.80 \\
4096 & 59.42 \\
8192 & 59.69 \\
16000 & \textbf{59.83} \\
32000 & 59.71 \\
\bottomrule
\end{tabular}
\caption{Accuracies for different values of $d$, the dimension of the compact bilinear feature. Models are trained on the VQA train split and tested on test-dev. Details in \Secref{sec:ablation}.}
\label{tab:comparedvalue}
\end{table}

\subsection{Ablation Results}
\label{sec:ablation}

\begin{table}[t]
\centering
\small
\begin{tabular}{@{}ll@{\ }l@{\ }l@{\ }l@{\ }l@{\ }l@{\ }l@{}}
\toprule
 \bf Method &  What &  Where &  When &  Who &  Why &  How & Avg \\ \cmidrule(lr){1-1}\cmidrule(lr){2-7}\cmidrule(lr){8-8}
Zhu et al. & 51.5 & 57.0 & 75.0 & 59.5 & 55.5 & 49.8 & 54.3 \\
 Concat+Att. & 47.8 & 56.9 & 74.1 & 62.3 & 52.7 & \bf 51.2 & 52.8 \\
 MCB+Att. & \bf 60.3 & \bf 70.4 & \bf 79.5 & \bf 69.2 & \bf 58.2 & 51.1 & \bf 62.2 \\
 \bottomrule
\end{tabular}
\caption{Multiple-choice QA tasks accuracy (\%) on Visual7W test set.}
\label{tab:Visual7W}
\end{table}

\begin{table*}[t]
\center
\begin{tabular}{lcccccccccc}
\toprule
  & \multicolumn{5}{c}{Test-dev} & \multicolumn{5}{c}{Test-standard} \\
\cmidrule(l){2-6}\cmidrule(l){7-11}
  & \multicolumn{4}{c}{Open Ended} & \multicolumn{1}{c}{MC} & \multicolumn{4}{c}{Open Ended} & \multicolumn{1}{c}{MC} \\
  \cmidrule(l){2-5}\cmidrule(l){6-6}\cmidrule(l){7-10}\cmidrule(l){11-11}
  & Y/N & No. & Other  & All & All & Y/N & No. & Other  & All & All \\
  \cmidrule(r){1-1}  \cmidrule(l){2-5}\cmidrule(l){6-6}\cmidrule(l){7-10}\cmidrule(l){11-11}
  MCB & 81.2 & 35.1 & 49.3 & 60.8 & 65.4 & - & - & - & - & - \\  %
  MCB + Genome   & 81.7 & 36.6 & 51.5 & 62.3 & 66.4 & - & - & - & - & - \\  %
  MCB + Att. & 82.2 & 37.7 & 54.8 & 64.2 & 68.6 & - & - & -  & - & - \\
  MCB + Att. + GloVe & 82.5 & 37.6 & 55.6 & 64.7 & 69.1 & - & - & - & - & - \\ %
  MCB + Att. + Genome & 81.7 & 38.2 & 57.0 & 65.1 & 69.5 & - & - & - & - & - \\ %
  MCB + Att. + GloVe + Genome & 82.3 & 37.2 & 57.4 & 65.4 & 69.9 & - & - & - & - & - \\ %
  Ensemble of 7 Att. models & \bf 83.4 & \bf 39.8 & \bf 58.5 & \bf 66.7 & \bf 70.2 & \bf 83.2 & \bf 39.5 & \bf 58.0 & \bf 66.5 & \bf 70.1 \\ %
    \cmidrule(r){1-1}  \cmidrule(l){2-5}\cmidrule(l){6-6}\cmidrule(l){7-10}\cmidrule(l){11-11}
    Naver Labs (challenge 2nd) & 83.5 & 39.8 & 54.8 & 64.9 & 69.4 & 83.3 & 38.7 & 54.6 & 64.8 & 69.3 \\
   HieCoAtt \cite{lu2016hiecoatt} & 79.7 & 38.7 & 51.7 & 61.8 & 65.8 & - & - & - & 62.1 & 66.1 \\
  DMN+ \cite{xiong16dynamic} & 80.5 & 36.8 & 48.3 & 60.3 & -& -&- &- & 60.4 & - \\
  FDA \cite{ilievski2016fda} & 81.1 & 36.2 & 45.8 & 59.2 & -& - & - & - & 59.5 & - \\
  D-NMN \cite{andreas16naacl} & 81.1 & 38.6 & 45.5 & 59.4 & - & - & - & - & 59.4 & -\\
  AMA \cite{wu16cvpr} & 81.0 & 38.4 & 45.2 & 59.2 & -& 81.1 & 37.1 & 45.8 & 59.4 & - \\
  SAN \cite{yang2015stacked} & 79.3 & 36.6 & 46.1 & 58.7 & -& - & - & - & 58.9 & - \\
NMN \cite{andreas16cvpr} & 81.2 & 38.0 & 44.0 & 58.6 & - & 81.2 & 37.7 & 44.0 & 58.7 & -  \\
AYN \cite{malinowski2016ask} & 78.4 & 36.4 &46.3 &58.4  & -& 78.2&36.3&46.3&58.4 & -\\
    SMem \cite{xu2015ask} & 80.9 & 37.3 & 43.1 & 58.0 & -& 80.9 & 37.5 & 43.5 & 58.2 & - \\
  VQA team \cite{antol2015vqa} & 80.5 & 36.8 & 43.1 & 57.8 & 62.7 & 80.6 & 36.5 & 43.7 & 58.2 & 63.1 \\
  DPPnet \cite{noh2015images} & 80.7 & 37.2 & 41.7 & 57.2 & -& 80.3 & 36.9 & 42.2 & 57.4 & -\\
  iBOWIMG \cite{zhou2015simple}  & 76.5 & 35.0 & 42.6 & 55.7 & -& 76.8 & 35.0 & 42.6 & 55.9 & 62.0 \\
\bottomrule
\end{tabular}
\caption{Open-ended and multiple-choice (MC) results on VQA test set (trained on train+val set) compared with state-of-the-art: accuracy in \%. See \Secref{sec:eval:stateoftheart}.}
\label{tab:vqa_state_of_art}
\vspace{-0.6cm}
\end{table*}

We compare the performance of non-bilinear and bilinear pooling methods in Table \ref{tab:comparefusion}. We see that MCB pooling outperforms all non-bilinear pooling methods, such as eltwise sum, concatenation, and eltwise product. 

One could argue that the compact bilinear method simply has more parameters than the non-bilinear pooling methods, which contributes to its performance. We compensated for this by stacking fully connected layers (with 4096 units per layer, ReLU activation, and dropout) after the non-bilinear pooling methods to increase their number of parameters. However, even with similar parameter budgets, non-bilinear methods could not achieve the same accuracy as the MCB method. For example, the ``Concatenation + FC + FC'' pooling method has approximately $ 4096^2 + 4096^2 + 4096\times 3000 \approx 46$ million parameters, which matches the 48 million parameters available in MCB with $d=16000$. However, the performance of the ``Concatenation + FC + FC'' method is only 57.10\% compared to MCB's 59.83\%.

Section 2 in Table \ref{tab:comparefusion} also shows that compact bilinear pooling has no impact on accuracy compared to full bilinear pooling. Section 3 in Table \ref{tab:comparefusion} demonstrates that the MCB brings improvements regardless of the image CNN used. We primarily use ResNet-152 in this paper, but MCB also improves performance if VGG-19 is used. Section 4 in Table \ref{tab:comparefusion} shows that our soft attention model works best with MCB pooling. In fact, attending to the Concatenation + FC layer has the same performance as not using attention at all, while attending to the MCB layer improves performance by 2.67 points.

Table \ref{tab:comparedvalue} compares different values of $d$, the output dimensionality of the multimodal compact bilinear feature. Approximating the bilinear feature with a 16,000-D vector yields the highest accuracy. 

We also evaluated models with multiple attention maps or channels. One attenion map achieves 64.67\%, two
 65.08\% and four 64.24\% accuracy (trained on train+val). Visual inspection of the generated attention maps reveals that an ensembling or smoothing effect occurs when using multiple maps. 

Table \ref{tab:Visual7W} presents results for the Visual7W multiple-choice QA task. The MCB with attention model outperforms the previous state-of-the-art by 7.9 points overall and performs better in almost every category. %

\subsection{Comparison to State-of-the-Art}
\label{sec:eval:stateoftheart}
Table \ref{tab:vqa_state_of_art} compares our approach with the state-of-the-art on VQA test set. Our best single model uses MCB pooling with two attention maps. Additionally, we augment our training data with images and QA pairs from the Visual Genome dataset. We also concatenate the learned word embedding with pretrained GloVe vectors \cite{pennington2014glove}.

Each model in our ensemble of 7 models uses MCB with attention. Some of the models were trained with data from Visual Genome, and some were trained with two attention maps. This ensemble is 1.8 points above the next best approach on the VQA open-ended task and 0.8 points above the next best approach on the multiple-choice task (on Test-dev). Even without ensembles, our ``MCB + Genome + Att. + GloVe'' model still outperforms the next best result by 0.5 points, with an accuracy of 65.4\% versus 64.9\% on the open-ended task (on Test-dev).

\section{Evaluation on Visual Grounding}

\begin{table}
\centering
\small
\begin{tabular}{lc}
\toprule
\bf Method & \bf Accuracy, \% \\ 
\twocmidrule
\newcite{plummer15iccv} & 27.42 \\
\newcite{hu16cvpr} & 27.80 \\
\newcite{plummer16arxiv}\footnotemark & 43.84 \\
\newcite{wang2016cvpr} & 43.89 \\
\newcite{rohrbach16arxiv} & 47.81 \\
\twocmidrule
Concatenation & 46.50 \\
Element-wise Product & 47.41 \\
Element-wise Product + Conv & 47.86 \\
MCB & \bf 48.69 \\
\bottomrule
\end{tabular}
\caption{Grounding accuracy on Flickr30k Entities dataset.}
\label{tab:Flickr30kEntities}
\end{table}
\footnotetext{\newcite{plummer16arxiv} achieve higher accuracy of 50.89\% when taking into account box size and color. We believe our approach would also benefit from such additional features.}

\begin{table}
\centering
\begin{tabular}{lc}
\toprule
 \bf Method & \bf Accuracy, \% \\ 
\twocmidrule
\newcite{hu16cvpr} & 17.93 \\
\newcite{rohrbach16arxiv} & 26.93 \\
\twocmidrule
Concatenation & 25.48 \\
Element-wise Product & 27.80 \\
Element-wise Product + Conv & 27.98 \\
MCB & \bf 28.91 \\
\bottomrule
\end{tabular}
\caption{Grounding accuracy on ReferItGame dataset.}
\label{tab:ReferItGame}
\end{table}

\begin{figure*}
\center
\begin{tabularx}{\textwidth}{XXXX}
\includegraphics[clip=true,width=0.22\textwidth,height=0.2\textheight,keepaspectratio]{fig/original_2.png} & 
\includegraphics[clip=true,width=0.22\textwidth,height=0.2\textheight,keepaspectratio]{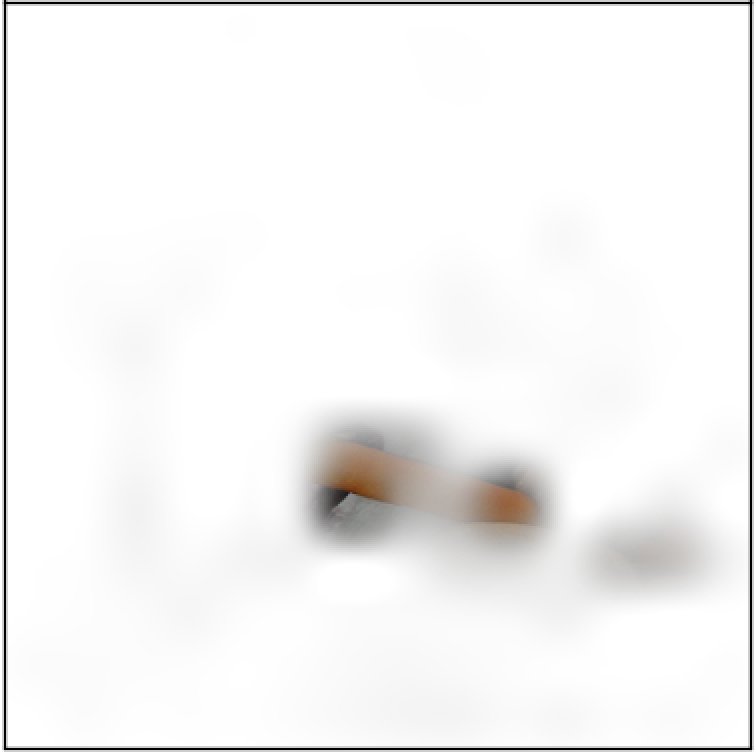} &
\includegraphics[clip=true,width=0.22\textwidth,height=0.2\textheight,keepaspectratio]{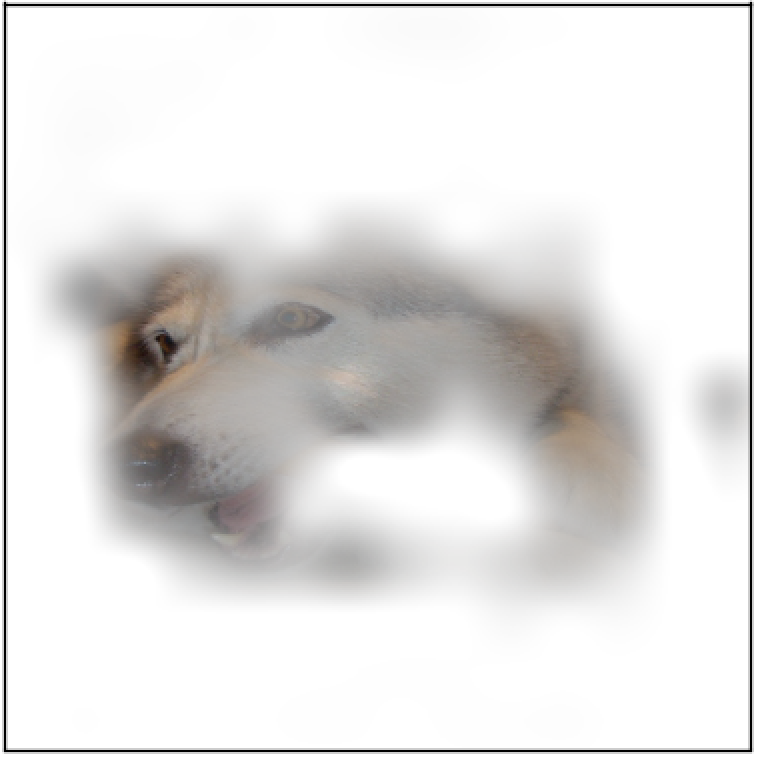} &
\includegraphics[clip=true,width=0.22\textwidth,height=0.2\textheight,keepaspectratio]{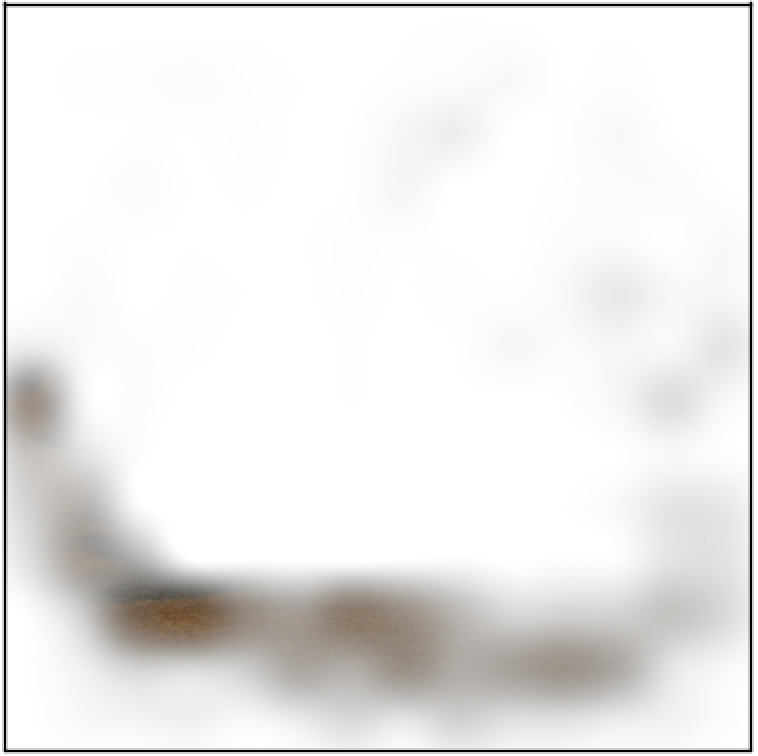} \\
&
\small{What vegetable is the dog chewing on? \newline {\color{blue} MCB: carrot} \newline {\color{ForestGreen} GT: carrot}} &
\small{What kind of dog is this? \newline {\color{blue} MCB: husky} \newline {\color{ForestGreen} GT: husky}} &
\small{What kind of flooring does the room have? \newline {\color{blue} MCB: carpet} \newline {\color{ForestGreen} GT: carpet}} \\
\includegraphics[clip=true,width=0.22\textwidth,height=0.2\textheight,keepaspectratio]{fig/original_1.png} & 
\includegraphics[clip=true,width=0.22\textwidth,height=0.2\textheight,keepaspectratio]{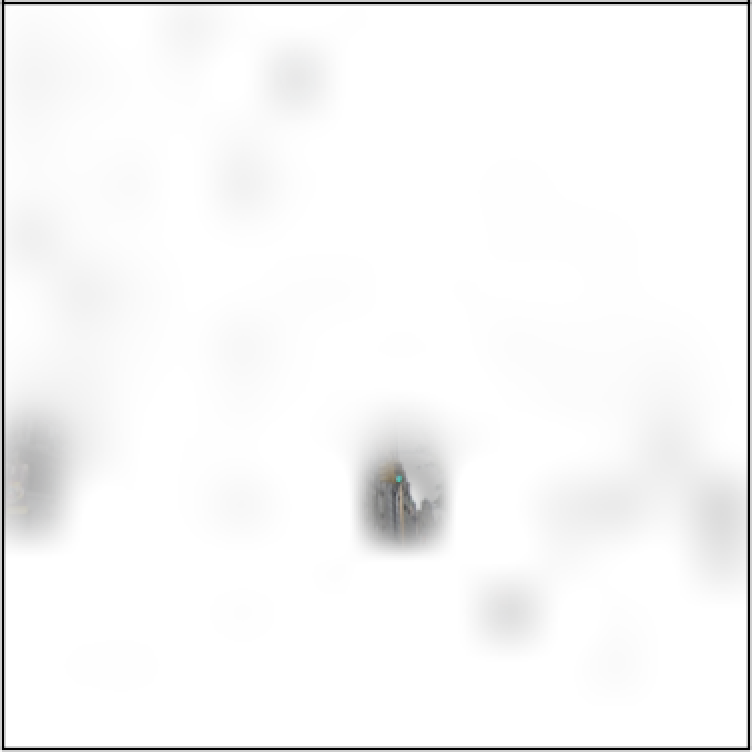} &
\includegraphics[clip=true,width=0.22\textwidth,height=0.2\textheight,keepaspectratio]{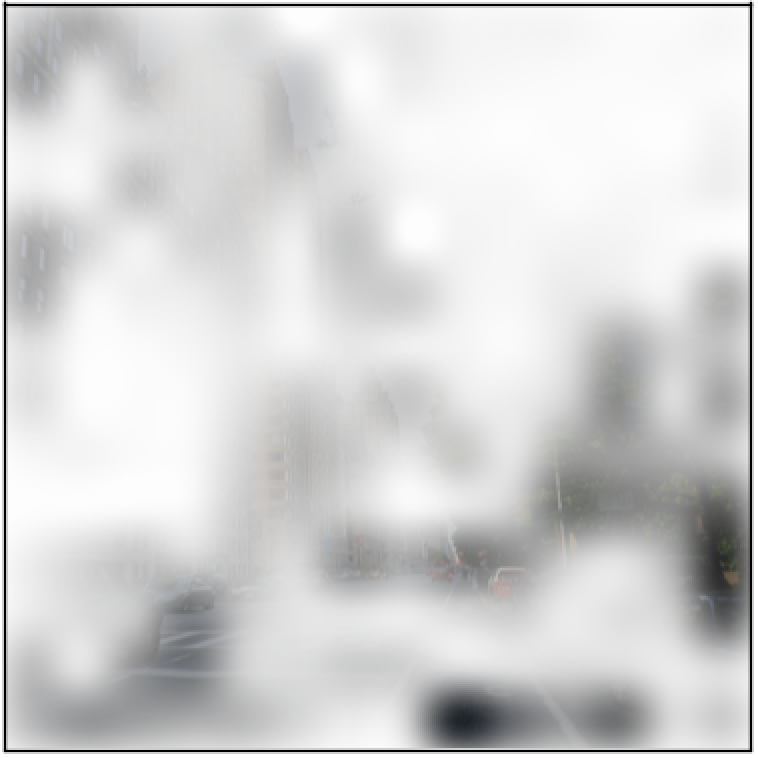} &
\includegraphics[clip=true,width=0.22\textwidth,height=0.2\textheight,keepaspectratio]{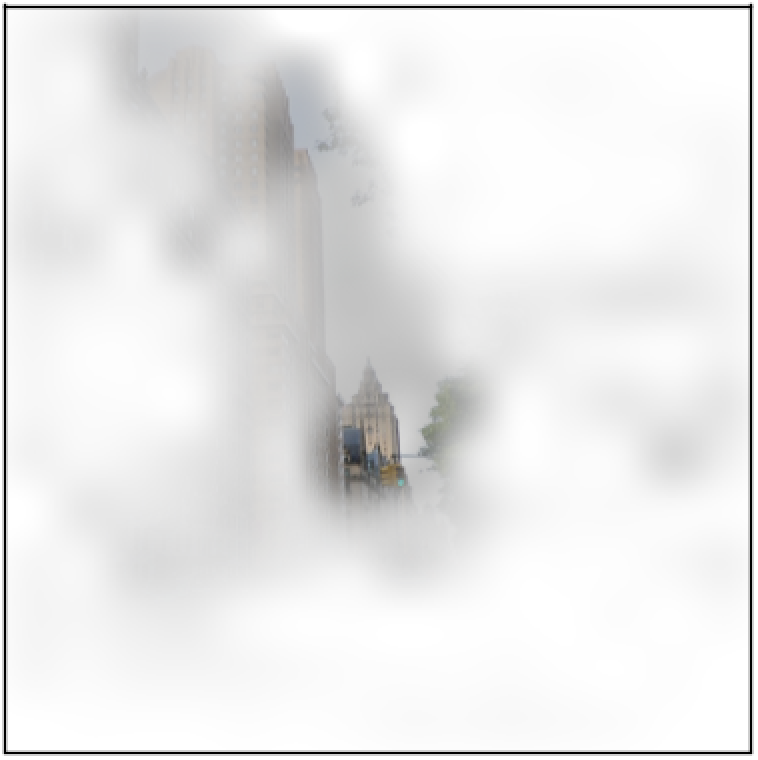} \\
&
\small{What color is the traffic light? \newline {\color{blue} MCB: green} \newline {\color{ForestGreen} GT: green}} &
\small{Is this an urban area? \newline {\color{blue} MCB: yes} \newline {\color{ForestGreen} GT: yes}} &
\small{Where are the buildings? \newline {\color{blue} MCB: in background} \newline {\color{ForestGreen} GT: on left}} \\
\hline
MCB&Eltwise Product + Conv&MCB&Eltwise Product + Conv\\
\includegraphics[clip=true,width=0.22\textwidth,keepaspectratio]{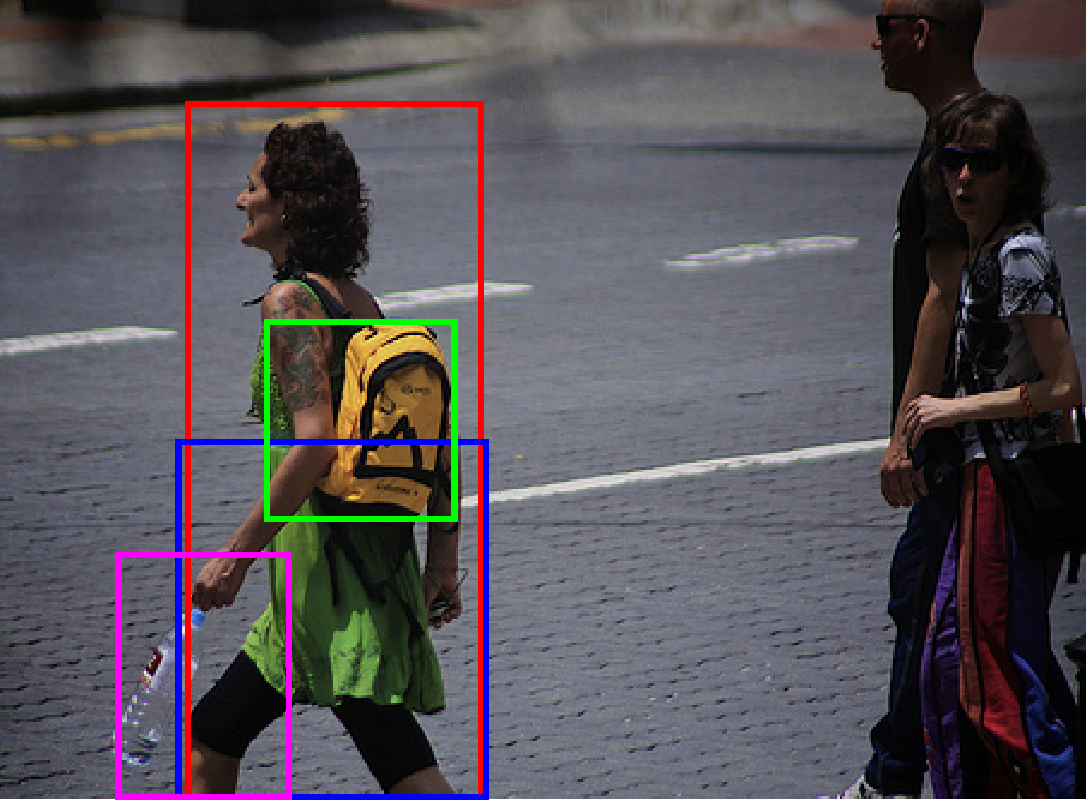} &
\includegraphics[clip=true,width=0.22\textwidth,keepaspectratio]{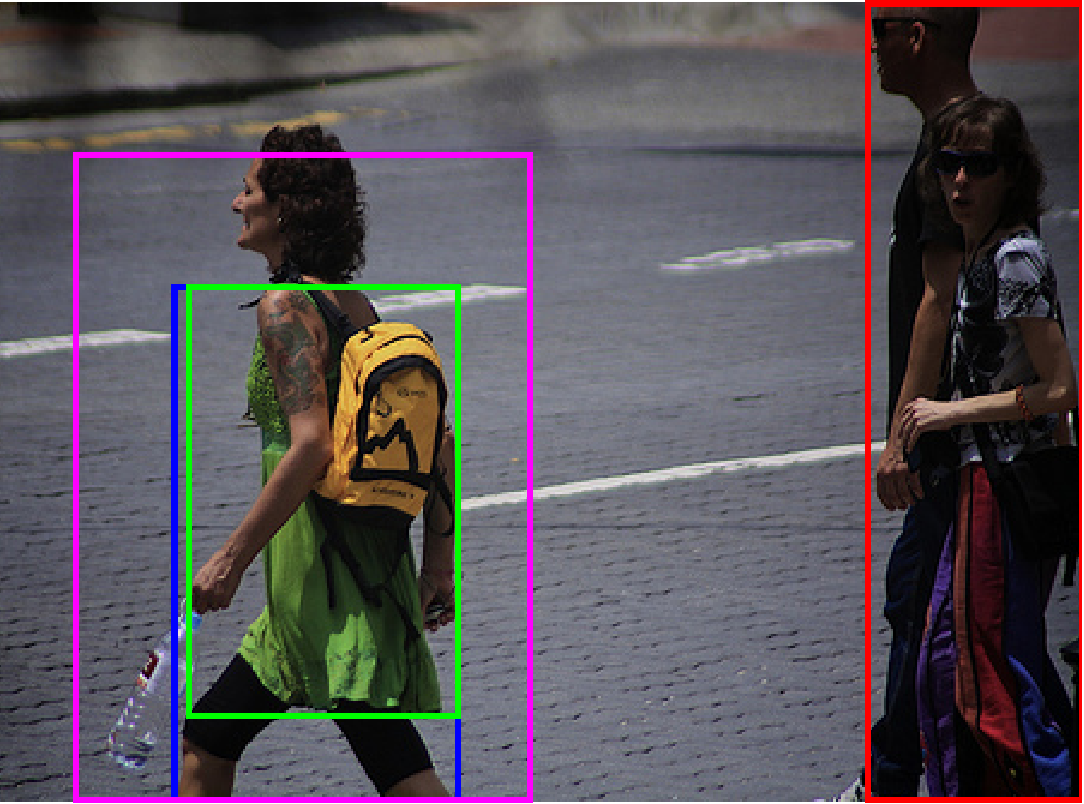} &
\includegraphics[clip=true,width=0.22\textwidth,keepaspectratio]{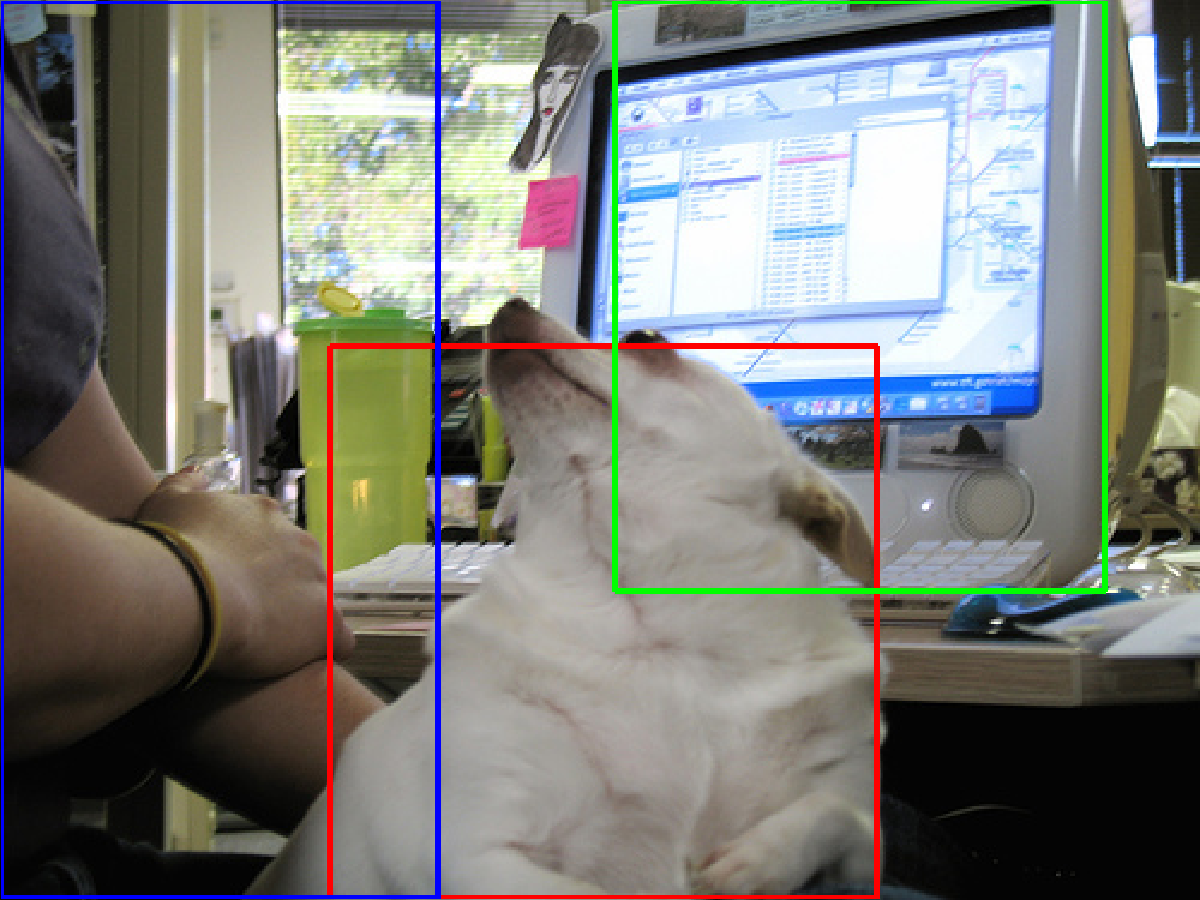} &
\includegraphics[clip=true,width=0.22\textwidth,keepaspectratio]{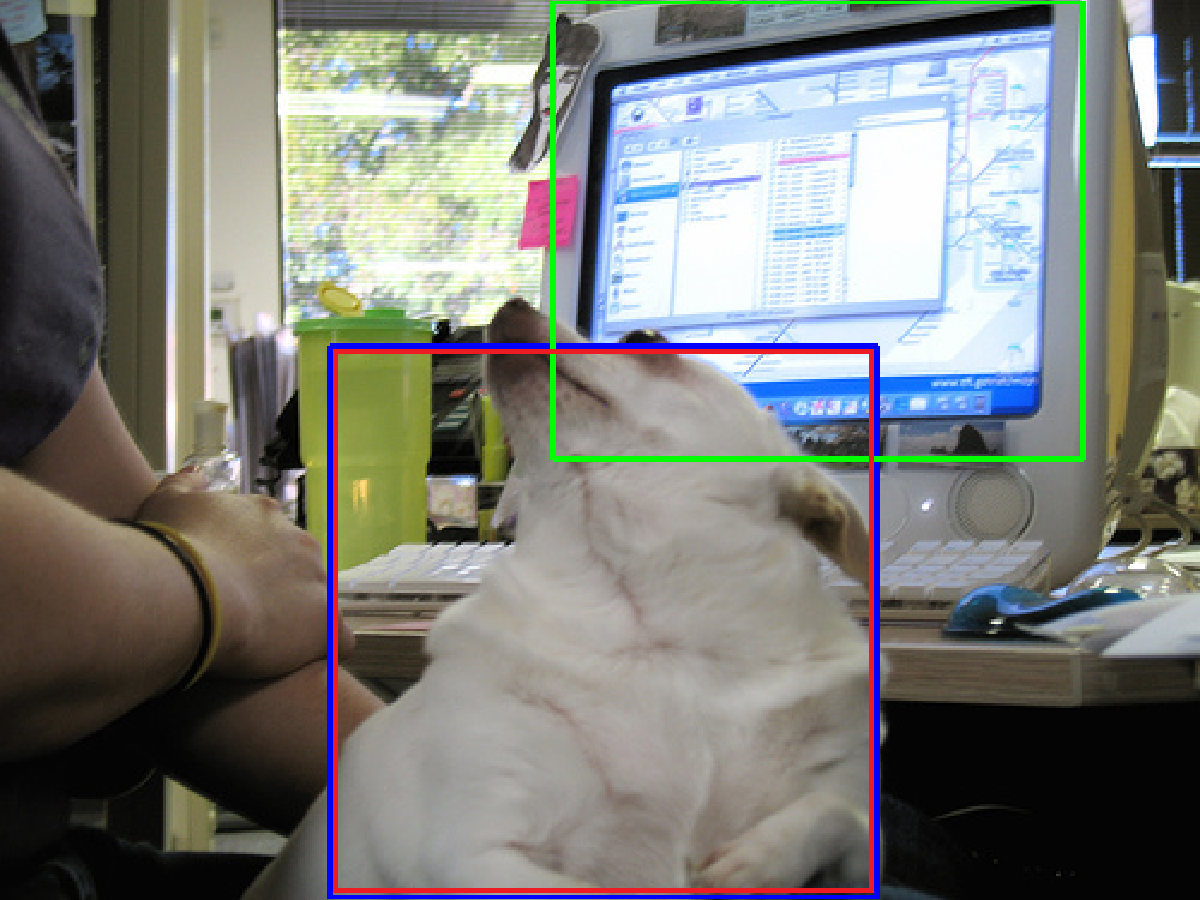} \\
\multicolumn{2}{p{7.7cm}}{\vspace{-0.3cm}\small{{\color{red} A tattooed woman} with {\color{blue} a green dress} and {\color{ForestGreen} yellow backpack} holding {\color{magenta} a water bottle} is walking across the street.}} &
\multicolumn{2}{p{7.7cm}}{\small{\vspace{-0.3cm}{\color{red} A dog} distracts {\color{blue} his owner} from working at {\color{ForestGreen} her computer}.}}
\end{tabularx}
\caption{Top: predicted answers and attention maps from MCB model on VQA images. Bottom: predicted grounding from MCB model (left) and Eltwise Product + Conv model (right) on Flickr30k Entities images.}
\label{fig:qualitative_att}
\end{figure*}

\subsection{Datasets}
We evaluate our visual grounding approach on two datasets. The first is Flickr30k Entities \cite{plummer15iccv} which consists of 31K images from Flickr30k dataset \cite{flickr30k} with 244K phrases localized with bounding boxes. We follow the experimental setup of \newcite{rohrbach16arxiv}, \eg we use the same Selective Search \cite{uijlings2013selective} object proposals and the Fast R-CNN \cite{girshick2015fast} fine-tuned VGG16 features \cite{simonyan2014very}. The second dataset is ReferItGame \cite{kazemzadeh14emnlp}, which contains 20K images from IAPR TC-12 dataset \cite{grubinger2006iapr} with segmented regions from SAIAPR-12 dataset \cite{escalante2010segmented} and 120K associated natural language referring expressions. For ReferItGame we follow the experimental setup of \newcite{hu16cvpr} and rely on their ground-truth bounding boxes extracted around the segmentation masks. We use the Edge Box \cite{zitnick2014eccv} object proposals and visual features (VGG16 combined with the spatial features, which encode bounding box relative position) from \newcite{hu16cvpr}.

\subsection{Experimental Setup}
In all experiments we use Adam solver \cite{kingma2014adam} with learning rate $\epsilon = 0.0001$. The embedding size is $500$ both for visual and language embeddings. We use $d=2048$ in the MCB pooling, which we found to work best for the visual grounding task. The accuracy is measured as percentage of query phrases which have been localized correctly. The phrase is localized correctly if the predicted bounding box overlaps with the ground-truth bounding box by more than $50\%$ intersection over union (IOU).

\subsection{Results}
Tables \ref{tab:Flickr30kEntities} and \ref{tab:ReferItGame} summarize our results in the visual grounding task. We present multiple ablations of our proposed architecture. First, we replace the MCB with simple concatenation of the embedded visual feature and the embedded phrase, resulting in $46.5\%$ on the Flickr30k Entities and $25.48\%$ on the ReferItGame datasets. The results can be improved by replacing the concatenation with the element-wise product of both embedded features ($47.41\%$ and $27.80\%$). We can further slightly increase the performance by introducing additional 2048-D convolution after the element-wise product ($47.86\%$ and $27.98\%$). However, even with fewer parameters, our MCB pooling significantly improves over this baseline on both datasets, reaching state-of-the-art accuracy of $48.69\%$ on Flickr30k Entities and $28.91\%$ on ReferItGame dataset. Figure \ref{fig:qualitative_att} (bottom) shows examples of improved phrase localization.

\section{Conclusion}
We propose the Multimodal Compact Bilinear Pooling (MCB) to combine visual and text representations. For visual question answering, our architecture with attention and multiple MCBs gives significant improvements on two VQA datasets compared to state-of-the-art. In the visual grounding task, introducing MCB pooling leads to improved phrase localization accuracy, indicating better interaction between query phrase representations and visual representations of proposal bounding boxes. The code to replicate our experiments is available at \url{https://github.com/akirafukui/vqa-mcb}.

\section*{Acknowledgments}
We would like to thank Yang Gao and Oscar Beijbom for helpful discussions about Compact Bilinear Pooling.
This work was supported by DARPA, AFRL, DoD MURI award N000141110688, NSF awards IIS-1427425 and IIS-1212798, and the Berkeley Artificial Intelligence Research (BAIR) Lab.

\bibliographystyle{emnlp2016}
\bibliography{biblioLong,mcbp16emnlp,rohrbach}

\end{document}